\documentclass[preprint,12pt]{elsarticle}

\usepackage{amssymb}
\usepackage{lineno}
\usepackage{amsmath}

\usepackage{algorithm}
\usepackage{algpseudocode}
\usepackage{geometry}
\usepackage{float}
\usepackage{import}
\usepackage{subfigure}
\usepackage{bm}
\usepackage{setspace}

\newtheorem{theorem}{Theorem}
\usepackage{xcolor}
\journal{Journal Name}
\RequirePackage[normalem]{ulem} %DIF PREAMBLE
\RequirePackage{color}\definecolor{RED}{rgb}{1,0,0}\definecolor{BLUE}{rgb}{0,0,1} %DIF PREAMBLE
\providecommand{\DIFdelbegin}{} %DIF PREAMBLE
\providecommand{\DIFdelend}{} %DIF PREAMBLE
\RequirePackage{listings} %DIF PREAMBLE
\RequirePackage{color} %DIF PREAMBLE
\lstdefinelanguage{DIFcode}{ %DIF PREAMBLE
  moredelim=[il][\color{red}\sout]{\%DIF\ <\ }, %DIF PREAMBLE
  moredelim=[il][\color{blue}\uwave]{\%DIF\ >\ } %DIF PREAMBLE
} %DIF PREAMBLE
\lstdefinestyle{DIFverbatimstyle}{ %DIF PREAMBLE
	language=DIFcode, %DIF PREAMBLE
	basicstyle=\ttfamily, %DIF PREAMBLE
	columns=fullflexible, %DIF PREAMBLE
	keepspaces=true %DIF PREAMBLE
} %DIF PREAMBLE
\lstnewenvironment{DIFverbatim}{\lstset{style=DIFverbatimstyle}}{} %DIF PREAMBLE
\lstnewenvironment{DIFverbatim*}{\lstset{style=DIFverbatimstyle,showspaces=true}}{} %DIF PREAMBLE
\begin{document}

	\begin{frontmatter}



		\title{Semi-Supervised Disentangled Framework for\\ Transferable Named Entity Recognition}


		\author[Address1,Address3]{Zhifeng Hao}

		\author[Address1]{Di Lv}

		\author[Address1]{Zijian Li}

		\author[Address1]{ Ruichu Cai*}
		\ead{cairuichu@gmail.com}
		\cortext[cor1]{Corresponding author}

		\author[Address1]{Wen Wen}
		\author[Address1]{Boyan Xu}

		\address[Address1]{School of Computer Science, Guangdong University of Technology, Guangzhou, China}
%		\address[Address2]{Beijing Key Laboratory of Intelligent Telecommunications Software and Multimedia, Beijing University of Posts and Telecommunications, Beijing, China}
		\address[Address3]{School of Mathematics and Big Data, Foshan University, Guangzhou, China}

		\begin{abstract}
			\DIFdelbegin %DIFDELCMD < \thispagestyle{empty}\pagestyle{empty} 
%DIFDELCMD < 			%%%
\DIFdelend %% Text of abstract
			Named entity recognition (NER) for identifying proper nouns in unstructured text is one of the most important and fundamental tasks in natural language processing. However, despite the widespread use of NER models, they still require  a large-scale labeled data set, which incurs a heavy burden due to manual annotation.
			Domain adaptation is one of the most promising solutions to this problem, where rich labeled data from the relevant source domain are utilized to strengthen the generalizability of a model based on the target domain. However, the mainstream cross-domain NER models are still affected by the following two challenges
			(1) Extracting domain-invariant information such as syntactic information for cross-domain transfer. 
			(2) Integrating domain-specific information such as semantic information into the model to improve the performance of NER.
			In this study, we present a semi-supervised framework for transferable NER, which disentangles the domain-invariant latent variables and domain-specific latent variables.
			In the proposed framework, the domain-specific information is integrated with the domain-specific latent variables by using a domain predictor. The domain-specific and  domain-invariant latent variables are disentangled using three mutual information regularization terms, i.e., maximizing the  mutual information between the domain-specific latent variables and the original embedding, maximizing the mutual information between the domain-invariant latent variables and the original embedding, and minimizing the mutual information between the domain-specific and domain-invariant latent variables.
			Extensive experiments demonstrated that our model can obtain state-of-the-art performance with  cross-domain and cross-lingual NER benchmark data sets.
		\end{abstract}		
		\begin{keyword}
			Named Entity Recognition \sep Semi-supervised Learning \sep Transfer Learning \sep Disentanglement
		\end{keyword}

	\end{frontmatter}

	%DIF < \linenumbers

	%% -------------------------

	%%--------------

	%------------------------------------------------
	\section{Introduction} % The \section*{} command stops section numbering
	
	{N}{amed} entity recognition(NER) is a standard natural language processing (NLP) task for identifying and classifying expressions with special meanings in unstructured text \cite{konkol2014named}.
	In recent years, NER approaches based on bidirectional long short-term memory (BiLSTM) and conditional random fields (CRF) have achieved excellent performance \cite{lample2016neural,cao2018adversarial}. However, these methods are domain specific and they cannot be readily generalized to data sets from other domains, mainly due to the excessive cost of manually constructing the required high-quality and large-scale data sets.

	Domain adaptation \cite{pan2009survey,ganin2015unsupervised,tzeng2015simultaneous,long2015learning,ijcai2019-285} aims to exploit abundant labeled data in the source domain to improve the performance in the target domains, and thus it can alleviate the restriction on NER caused by the limited availability of labeled data in the  target domain. Most of the existing domain-adaptation approaches were designed for unsupervised scenarios where both the source and target domains share the same-named output space. However, these conventional domain adaptation approaches are not suitable in cases where the named entities have a different source domain to the target domain.

	In order to enable transfer learning in the settings with different source domain and target domain entity spaces, a few labeled target-domain samples should be incorporated into the training data set, which is referred to as semi-supervised domain adaptation for NER \cite{peters2017semi}.
	There are two main research areas for transferable NER. The first category comprises simple and straightforward methods. For example, Lee et al. used  labeled target domain samples to fine tune a model initialized based on a source domain data set \cite{lee2018transfer}. The other types of methods such as that proposed by Yang et al. are based on the idea of multi-task learning and developing a model that contains a shared feature extractor and a domain-specific CRF layer for the source and target domains, respectively \cite{yang2016multi, yangtransfer}.

	However, the methods mentioned above have the following limitations.
	(1) The domain-invariant information is not explicitly extracted. For example, fine tuning-based methods \cite{lee2018transfer} may not work very well when the gap between the source domain and target domain distribution is excessively large because a few target-domain samples may lead to overfitting even when  domain-specific information is considered.
	(2) The domain-specific information is usually ignored. For example, multi-task-based methods \cite{yangtransfer,yang2016multi} implicitly assume  that a shared feature extractor can generate the domain-invariant information. However, they do not perform well at recognizing domain-specific name entities because the domain-specific information is not well integrated.

	\begin{figure}[H]
		\centering
		\includegraphics[width=\columnwidth]{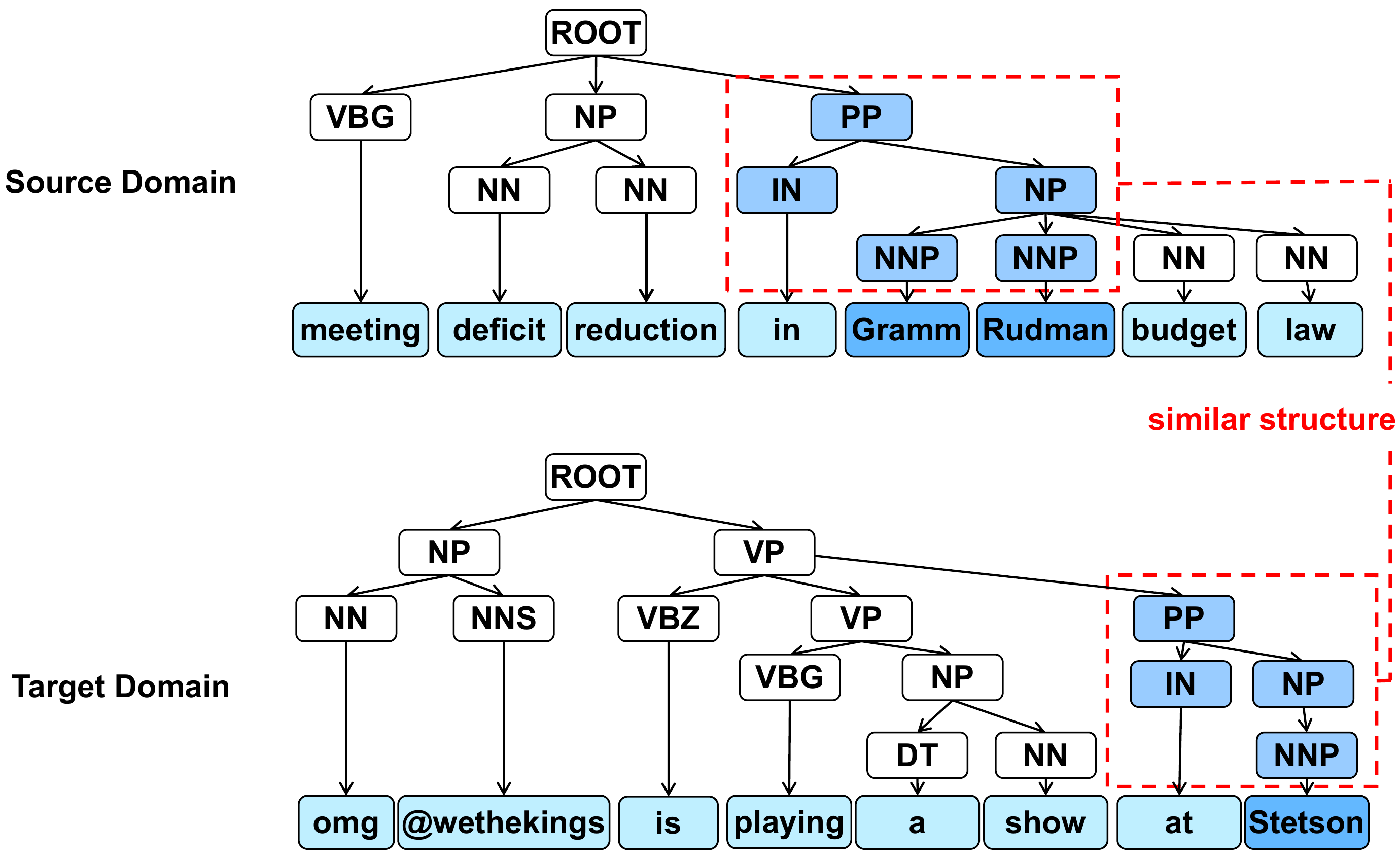}
		\caption{Illustrations of syntactic structure trees from source and target domains. The upper and lower syntactic structure trees were parsed using  sentences from Newswire and social media domains. ``Gramm" and ``Rudman" are related to law in the source domain, and the tag ``Stetson" is related to  facility in the target domain. The topics of different domains are different and we treated them as domain-specific information. The red boxes denote the similar substructure from different domains and they indicate domain-invariant information.}
		\label{fig:motivation}
	\end{figure}
	% 半监督迁移的NER问题在于如何同时利用领域特有和领域共同的知识
	To address these problems, it is necessary to find a solution that can extract and utilize both the domain-invariant and domain-specific information.
	Figure \ref{fig:motivation} illustrates an example of domain-invariant and domain-specific components in NER.
	In this example, the topic of the source domain (``Newswire'') is obviously different from that of the target domain (``social media''), which is domain-specific information. However, the syntactic substructures, which are important for locating the name entities, are similar in the two domains and they  can be considered as domain-invariant information.
	Hence, we can assume that each sample is controlled by two independent latent variables $\bm{z}$ and $\bm{v}$, which we denote as domain-specific and domain-invariant latent variables, respectively.
	Our aim is to disentangle these two latent variables.
	Cai et al. utilized an analogous technique for unsupervised domain adaptation by using two supervised signals \cite{ijcai2019-285}. In addition, Chen et al. employed paraphrase pair data sets in a subtle manner and learned sentence representations to disentangle the syntax and semantics of a sentence by incorporating the semantic and syntactic supervised signals \cite{chen2019multi}. However, it is still very challenging to disentangle these latent variables in cross-domain NER because it is difficult to obtain a data set with labels that indicate whether two sentences have similar substructures.

	In the present study, inspired by the disentangled representations of multiple explanatory factors used in previous research \mbox{%DIFAUXCMD
\cite{bengio2013representation,locatello2019challenging, dinh2014nice}}\hspace{0pt}%DIFAUXCMD
, we developed a semi-supervised disentangled (\textbf{SSD}) framework for transferable NER, which assumes that the domain-specific variables $\bm{z}$ are independent of the domain-invariant variables $\bm{v}$.
	In the proposed \textbf{SSD} framework, the domain-specific latent variables $\bm{z}$ and domain-invariant latent variables $\bm{v}$ are extracted, disentangled, and then simultaneously used to predict the named entities.
	In order to disentangle two latent variables with limited supervision of the signals, we first use a domain predictor to push the domain-specific information into $\bm{z}$, before then employing three types of mutual information regularization terms. In particular, we simultaneously maximize the mutual information between the domain-specific latent variables and the original  embedding, maximize the mutual information between the domain-invariant latent variables and the original embedding, and minimize the mutual information between the domain-specific and domain-invariant latent variables. Our \textbf{SSD} model estimates the mutual information by using neural networks \cite{belghazi2018mutual} and we optimize our \textbf{SSD} model in an iterative strategy, which guarantees the accuracy of the estimated mutual information.
	Extensive experiments demonstrated that \textbf{SSD} outperformed the state-of-the-art transferable NER methods based on cross-domain and cross-lingual standard benchmarks.

	The main contributions of our study are summarized as follows.
	\begin{itemize}
	\item We propose a semi-supervised framework for transferable NER by disentangling domain-invariant and domain-specific information.
	\item In the proposed framework, we employ three mutual information regularization terms to successfully achieve disentanglement with limited supervision of the signals.
	\item In the proposed framework, we utilize both the domain-invariant and domain-specific information to accurately recognize a named entity based on the target domain.
	\item Experimental studies demonstrated that our model obtained state-of-the-art performance with cross-domain and cross-lingual data sets.
	\end{itemize}

	The remainder of this paper is organized as follows. In Section \ref{related_work}, we review related research into NER, domain adaptation, domain adaptation in NLP, and disentanglement. In Section \ref{preliminary}, we define the problem of semi-supervised domain adaptation for NER and describe some preliminary techniques.
	In Section \ref{model}, we give the details of our \textbf{SSD} model.
	In Section \ref{experiment}, we present our experimental results based on standard benchmarks. In Section \ref{conclusion}, we give our conclusions and suggestions regarding future research.

\section{Related Work} \label{related_work}
%In this section, we first review the existing techniques on named entity recognition and domain adaptation. Then we give a a brief introduction about domain adaptation in natural language processing, especially in named entity recognition. Finally, we introduce the mainstream approaches of disentanglement and disentanglement in natural language processing.

\noindent\textbf{NER:} Automatic detection of named entities in free text is a fundamental task in information extraction. Many downstream tasks such as question answering \cite{duan2017question} and text summarization \cite{collins2017supervised} depend on the performance of NER. Traditional approaches to NER include CRF models \cite{lafferty2001conditional} and maximum entropy Markov models \cite{mccallum2000maximum}. In recent years, several deep learning-based NER methods have been proposed \mbox{%DIFAUXCMD
\cite{lample2016neural, cao2018adversarial, tran2017named, liu2018empower}}\hspace{0pt}%DIFAUXCMD
. These methods share a similar architecture, which employs different levels of embedding, BiLSTM for sequence modeling, and a CRF layer \cite{lafferty2001conditional} to predict labels. Due to the advantages of the neural network, little feature engineering is required to train a NER model. We employ the BiLSTM-CRF architecture as the backbone network for our \textbf{SSD} model.

\noindent\textbf{Domain Adaptation:} Domain adaptation \cite{pan2009survey,ganin2015unsupervised,tzeng2015simultaneous,long2015learning,ijcai2019-285} is a hot topic in machine learning. The mainstream methods applied in the unsupervised scenario aim to extract the domain-invariant features between domains. Maximum mean discrepancy \cite{ghifary2016scatter} is one of the most popular methods employed, which uses a geometrical measure that operates in the reproducing kernel Hilbert space. Another typical approach involves extracting the domain-invariant representation by introducing a gradient reversal layer \cite{ganin2015unsupervised} for domain alignment.
These conventional approaches are mainly designed for unsupervised domain adaptation, where it is assumed that the domain-invariant information plays an important role in decisions and that different domains share the same label space.
In the present study, we consider the problem of cross-domain NER where the label spaces of the source and target domains are different, so domain-specific information also plays an important role. Therefore, semi-supervised domain adaptation \cite{xiao2012semi} is employed.

\noindent\textbf{Domain Adaptation in NLP: }
Due to the excessive cost incurred to achieve the expected data quality and quantity, domain adaptation is also extremely important for many NLP tasks. %There are lots of NLP tasks that are combined with domain adaptation.
For example, Li et al. \cite{li2018hierarchical} simultaneously utilized both domain-specific and domain-shared sentiment words for sentiment classification. Hu et al. \cite{hu2019domain} proposed an unsupervised domain adaptation method for neural machine translation by constructing a pseudo-parallel in-domain corpus.
Recently, cross-domain NER has attracted widespread interest in the field of machine learning. Considering that some domain-invariant knowledge can be transferred from the source to the target domain, Lee et al. \cite{lee2018transfer} directly used the target data set to fine tune a model initialized with the source data set. Based on  the idea of multi-task learning \cite{lee2018transfer}, Yang et al. \cite{yangtransfer} considered the source and target domains as different tasks and extracted the domain-invariant information by multi-task learning. However, these multi-task-based methods \cite{lin2018neural,yangtransfer} ignore the differences in the output space across domains, which may result in negative transfer.
Lin et al. \cite{lin2018neural} solved this problem by appending an input adaptation layer after the word embedding layer and an output adaptation layer before the classifier.
However, domain-specific information in the data sets is also important but the aforementioned methods do not use it explicitly.

\noindent\textbf{Disentanglement:} Disentangled representation \cite{bengio2013representation} means that a change in one dimension corresponds to a change in one factor of variation, but the other factor is invariant. Several interesting studies have investigated disentangled representations for computer vision tasks based on a variational autoencoder \cite{kingma2013auto, kim2018disentangling, higgins2016beta, locatello2019challenging}.
Cai et al. \cite{ijcai2019-285} proposed a disentangled semantic representation model for unsupervised domain adaptation. For NLP tasks, the highly related words comprise the disentanglement between the syntax and the semantics of a sentence. Chen et al. \cite{chen2019multi} proposed an approach to disentangle high-level information by skillfully utilizing the paraphrase pairs data set. In contrast to Chen et al. \cite{chen2019multi} who used semantic labels and syntactic labels to disentangle the semantic and syntactic structure information, our \textbf{SSD} framework only exploits the domain label that represents different semantic information to disentangle the domain-specific and domain-invariant information by using three mutual information regularization terms.

We propose an SSD model for transferable NER, which disentangles the domain-invariant and domain-specific information, and simultaneously uses both for recognizing named entities in the target domain.

\section{Preliminaries} \label{preliminary}
First, we define the problem of semi-supervised domain adaptation for NER, before provising a brief introduction to the basic model.

\subsection{Problem Definition}
Let $\bm{x}=[x_1, x_2, ..., x_L]$ be a sentence with $L$ words, $\bm{y}=[y_1, y_2,..., y_L]$ is the label sequence where $y_i \in E$, and $E$ is the named entity set.
Let $E_S$ and $E_T$ be the entity sets of the source domain and target domain, respectively.
Given the training data set $D=\{(\bm{x}_s,\bm{y}_s)\}_{s=1}^M\cup\{(\bm{x}_t,\bm{y}_t)\}_{t=1}^N$, where $\bm{y}_s\subseteq E_S, \bm{y}_t \subseteq E_T$ and $M \gg N$,
our objective is to devise a model that can learn from the training data set and then predict a label sequence for the test data set $D^*=\{\bm{x}_t^*\}_{t=1}^{N^*}$ in the target domain.

\subsection{Basic Model}
BiLSTM with a CRF layer \cite{lafferty2001conditional} and self-attention mechanism \cite{cao2018adversarial} is used as the basic model for transferable NER because of its significant advantages compared with the conventional approach \cite{lafferty2001conditional}. In the following, we present some details of BiLSTM with a CRF layer, the self-attention mechanism, and its application in the semi-supervised domain adaptation setting.

\subsubsection{\textbf{Input Embedding}} \label{input_embedding}
The first step in the model is to map the discrete words into the distributed representation. Given a sentence $\bm{x}=[x_1, x_2, ... x_L]$, we look up the embedding vector $\bm{w}_i$ from the pre-trained embedding matrix. The sensitivity of the spelling should be considered, so we also look up the character-level embedding vector $\bm{c}_{ij}$ in the character-level embedding matrix for each character, i.e., $\bm{c}_{ij}$ denotes the character-level embedding vector of the $j$-th letter in the $i$-th word. We then use a convolutional neural network and max pooling to extract the character-level representation $\bm{c}_i$ of the $i$-th word \cite{chiu2016named}. Formally, we define the character-level feature extraction process as follows:
\begin{equation}\label{input_embedding}
\begin{split}
\bm{c}_i = Pooling(Conv(\bm{c}_{i1}, \bm{c}_{i2}, ..., \bm{c}_{ij}, ..., \bm{c}_{iJ};\bm{\theta_c})),
\end{split}
\end{equation}
where $J$ represents the character number of word $x_i$; $Pooling$ and $Conv$ denote the max pooling and convolutional neural network, respectively; and $\bm{\theta_c}$ denotes the parameters of CNN. Then, the character-level representation $\bm{c}_i$ is concatenated with the word embedding $\bm{w}_i$ as follows:
\begin{equation}
\begin{split}
\widetilde{\bm{w}}_i &= \bm{w}_i \oplus \bm{c}_i,
\end{split}
\end{equation}where $\oplus$ is the concatenation operation and $\widetilde{\bm{w}}_i$ is the final input embedding of $\bm{x}_i$. For convenience, we define the aforementioned process as follows:
\begin{equation}\label{merge_embedding}
\begin{split}
\widetilde{\bm{w}}=G_e(x_i;\bm{\theta_c}).
\end{split}
\end{equation}
We obtained Equations (\ref{input_embedding})--(\ref{merge_embedding}) from the study by\cite{chiu2016named}.

\subsubsection{\textbf{BiLSTM for Sequence Modeling}} \label{bi-lstm}
Next, based on the study by \mbox{%DIFAUXCMD
\cite{chiu2016named}}\hspace{0pt}%DIFAUXCMD
, we present the basic features of BiLSTM and its usage in sequence modeling. First, we define:
\begin{equation}
\begin{split}\label{bilstm_equation}
\stackrel{\rightarrow}{\bm{h}_i}&=\stackrel{\longrightarrow}{LSTM}(\stackrel{\longrightarrow}{\bm{h}_{i-1}}, \bm{\widetilde{\bm{w}}}_i), \\
\stackrel{\leftarrow}{\bm{h}_i}&=\stackrel{\longleftarrow}{LSTM}(\stackrel{\longleftarrow}{\bm{h}_{i+1}}, \bm{\widetilde{w}}_i), \\
\bm{h}_i&=\stackrel{\rightarrow}{\bm{h}_i} \oplus \stackrel{\leftarrow}{\bm{h}_i},
\end{split}
\end{equation}
where $\stackrel{\rightarrow}{\bm{h}_i} \in \mathbb{R}^d$ and $\stackrel{\leftarrow}{\bm{h}_i} \in \mathbb{R}^d$ denote the hidden states of the forward and backward LSTM at the $i$-th time step, respectively. Formally, we define the aforementioned process as follows:
\begin{equation}\label{hidden_state}
\begin{split}
\bm{h}=\left[ \bm{h}_1, \bm{h}_2, ...\bm{h}_L \right] = G_r\left(\widetilde{\bm{w}};\bm{\theta_r}\right),
\end{split}
\end{equation}
where $\bm{h}$ represents all the hidden states of BiLSTM and $\bm{\theta_r}$ denotes the parameters of BiLSTM. We describe the BiLSTM sequence model according to \mbox{%DIFAUXCMD
\cite{chiu2016named} }\hspace{0pt}%DIFAUXCMD
by Equations (\ref{bilstm_equation})--(\ref{hidden_state}).
%And we further let $h=\left[ \bm{h}_1, \bm{h}_2, ...\bm{h}_L \right]$ as all the hidden states of bi-LSTM given $x$.

\subsubsection{\textbf{Self-Attention Mechanism}} \label{self_attention}
We utilize a multi-head self-attention mechanism to extract the dependencies among words in a sentence and capture the inner syntactic structure information in a similar manner to Cao et al.\cite{cao2018adversarial}.

The attention mechanism maps a query and a set of key--value pairs to an output. In the self-attention mechanism, the query ($\bm{Q} \in \mathbb{R}^{L \times 2d}$), key ($\bm{K} \in \mathbb{R}^{L \times 2d}$), and value ($\bm{V} \in \mathbb{R}^{L \times 2d}$) are actually the hidden states described in \ref{bi-lstm}. The first step of the multi-head attention mechanism involves linearly projecting the query, key, and value $\tau$ times by using different linear projections. The $t$-th projection is as follows:
\begin{equation}\label{attention_begin}
\begin{split}
head_t &= Attention\left(\bm{Q}\bm{W}_t^Q, \bm{K}\bm{W}_t^K, \bm{V}\bm{W}_t^V\right)\\
&=softmax\left( \frac{ \left(\bm{Q}\bm{W}_t^Q\right)\left( \bm{K}\bm{W}_t^K\right)^\mathsf{T} }{\sqrt{u}} \right) \bm{V}\bm{W}_t^V,
\end{split}
\end{equation}
where $\bm{W}_t^Q \in \mathbb{R}^{2d \times u}$, $\bm{W}_t^K  \in \mathbb{R}^{2d \times u}$, and $\bm{W}_t^V \in \mathbb{R}^{2d \times u}$ are trainable projection parameters and $t=1,2,3,\cdots, \tau$. These results are then concatenated and projected to generate the final representation $\widetilde{h_i}$, which is defined as follows:
\begin{equation}
\begin{split}
\widetilde{\bm{h}}_i = \left(head_1 \oplus head_2, ...\oplus head_t ...\oplus head_\tau\right)\bm{W}_o,
\end{split}
\end{equation}
where $\bm{W}_o$ are also trainable parameters. This process is described as follows:
\begin{equation}\label{attention_end}
\begin{split}
\widetilde{\bm{h}}&=G_a\left(\bm{h};\bm{\theta_a}\right),
%\widetilde{\bm{y}}&=G_y\left(\widetilde{\bm{h}}; \bm{\theta_y}\right),
\end{split}
\end{equation}
where $\bm{\theta_a}=\{\bm{W}_t^Q, \bm{W}_t^K, \bm{W}_t^V, \bm{W}_o \}$ denote the parameters of the self-attention mechanism. We obtained Equation (\ref{attention_begin})--(\ref{attention_end}) from the study by \cite{cao2018adversarial}.

\subsubsection{\textbf{CRF Layer for Label Prediction}}\label{CRF}
The CRF layer used in our framework is based on the previous study by \cite{lafferty2001conditional}. In particular, the probabilistic model of the CRF sequence defines a family of conditional probabilities $p(\bm{y}|\bm{\widetilde{h}};\bm{w}_y, \bm{b}_y)$ over all possible label sequences $\bm{y}$ given $\bm{\widetilde{h}}$, with
the following form:
\begin{equation}
\begin{split}
	%\mathcal{L}_y\left(\theta_c, \theta_r, \theta_a, \theta_y^S, \theta_y^T\right) =
	p(\bm{y}|\bm{\widetilde{h}};w, b) = \cfrac{\displaystyle\prod_{i = 1}^{|\bm{y}|}\psi_i({y}_{i-1},{y}_i,\bm{\widetilde{h}})}{\displaystyle\sum_{{\bm{y}}^{'}\in\mathcal{Y}(\bm{\widetilde{h}})}\displaystyle\prod_{i = 1}^{|{\bm{y}}^{'}|}\psi_i(y^{'}_{i-1},y^{'}_i,{\bm{\widetilde{h}}})}
	\label{loss_func}
\end{split}
\end{equation}
where $\psi_i(\bm{y}_{'},\bm{y}, \bm{\widetilde{h}}) =$ exp$(\bm{w}^T_{y',y}\widetilde{\bm{h}}_{i} + \bm{b}_{y',y})$ are potential functions, and $\bm{w}^T_{y',y}$ and $\bm{b}_{y',y}$ are the weight vector and bias corresponding to label pair $(\bm{y}^{'}, \bm{y})$, respectively. For convenience, we let $\theta_{\bm{y}} = \{\bm{w}, \bm{b}\}$.\par
%For a training set $\{(\bm{\widetilde{h}}_{i},y_{i})\}$, the logarithm of the likelihood is given by:
%\begin{equation}
%\begin{split}
%\mathcal{L}_{\bm{y}}(\bm{\theta_y}) = \displaystyle\sum_{i}p(\bm{y}|\bm{\widetilde{h}};\bm{\theta}_{\bm{y}})
%\label{loss_func}
%\end{split}
%\end{equation}
%Maximum likelihood training chooses parameters such that the log-likelihood $\mathcal{L}_{\bm{y}}(\bm{\theta})$ is maximized.
%Decoding is
Therefore, the CRF layer is used to search for the label sequence $\bm{y}^{*}$ with the highest conditional probability, as follows.
\begin{equation}
\begin{split}
\bm{y}^{*} = \arg\max_{\bm{y}\in\mathcal{Y}(\bm{\widetilde{h}})}p(\bm{y}|\bm{\widetilde{h}};\bm{\theta_y})
\label{loss_func}
\end{split}
\end{equation}
For a sequence CRF model, training and decoding can be solved efficiently using the Viterbi algorithm. Given a ground truth sequence $y$ and a predicted sequence $y^*$, the loss can be represented as $L_{CRF}\left(y^*, y\right)$.

\subsubsection{\textbf{Semi-Supervised Domain Adaptation Training Method}}
The source and target domain have different label sets, so the CRF layer mentioned in \ref{CRF} cannot share parameters across two domains, i.e., each domain learns a separate CRF layer. Therefore, we extend the CRF layer for label prediction and let $G_y^S\left(.;\bm{\theta_y^S}\right)$ and $G_y^T\left(.;\bm{\theta_y^T}\right)$ be the CRF layers for the source and target domains, respectively. For convenience, we let $\{ \bm{\theta_c}, \bm{\theta_r}, \bm{\theta_a}, \bm{\theta_y^S}, \bm{\theta_y^T}\} = \bm{\Theta}$ and they are trained by minimizing the following objective function.
\begin{equation}
	\begin{split}
	\mathcal{L}_y\left(\Theta\right) = \frac{1}{M}\mathcal{L}_{CRF}\left(y^*_S, y_S\right) + \frac{1}{N}\mathcal{L}_{CRF}\left(y^*_T, y_T\right)
	\end{split}
\end{equation}

In the next section, we introduce our SSD framework for cross-domain NER.

\section{MODEL} \label{model}

\begin{figure}[H]
	\centering
	\includegraphics[width=0.25\columnwidth]{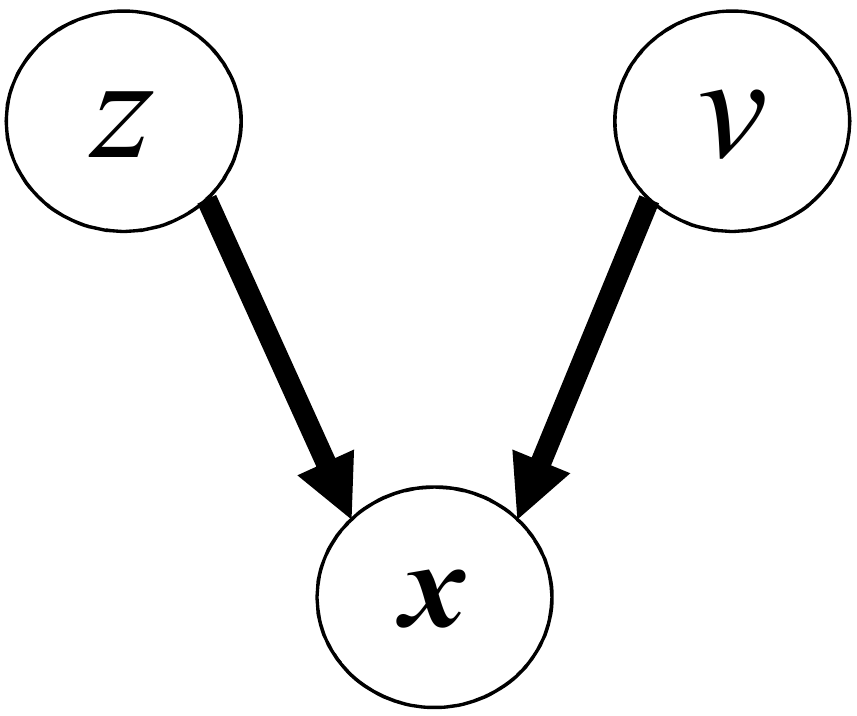}
	\caption{Causal model of the data generation process, which is controlled by the domain-specific latent variables $\bm{z}$ and domain-invariant latent variables $\bm{v}$.}
	\label{fig:generator}
\end{figure}

As illustrated in Fig. \ref{fig:motivation}, a sentence $\bm{x}$ can be generated from two independent latent variables: the domain-invariant variables $\bm{v}$ and the domain-specific variables $\bm{z}$.
Hence, the causal mechanism for the data generation process can be described as Fig. \ref{fig:generator}.
Intuitively, entities tend to be located in similar syntactic structures in sentences from two domains, thereby making $\bm{v}$ transferable.
In addition, the domain-specific variables, such as topics, are unique and definitely related to either the source or the target domain.

According to this observation, we need a model that can disentangle and utilize the domain-invariant latent variables and domain-specific latent variables. In the semi-supervised transferable NER, we assume that the domain-invariant latent variables contain the syntactic information and that the domain-specific latent variables contain the semantic information. Previous methods proposed for the disentanglement of semantic and syntactic information \mbox{%DIFAUXCMD
\cite{chen2019multi} }\hspace{0pt}%DIFAUXCMD
require two types of labels: semantic similarity labels and syntactic structure similarity labels. However, the syntactic structure similarity labels are difficult to obtain. Thus, in order to address this problem, we propose the \textbf{SSD} for transferable NER, which disentangles these two variables via domain label supervision and the three mutual information regularization terms.

\begin{figure}
	\centering
	\includegraphics[width=0.9\columnwidth]{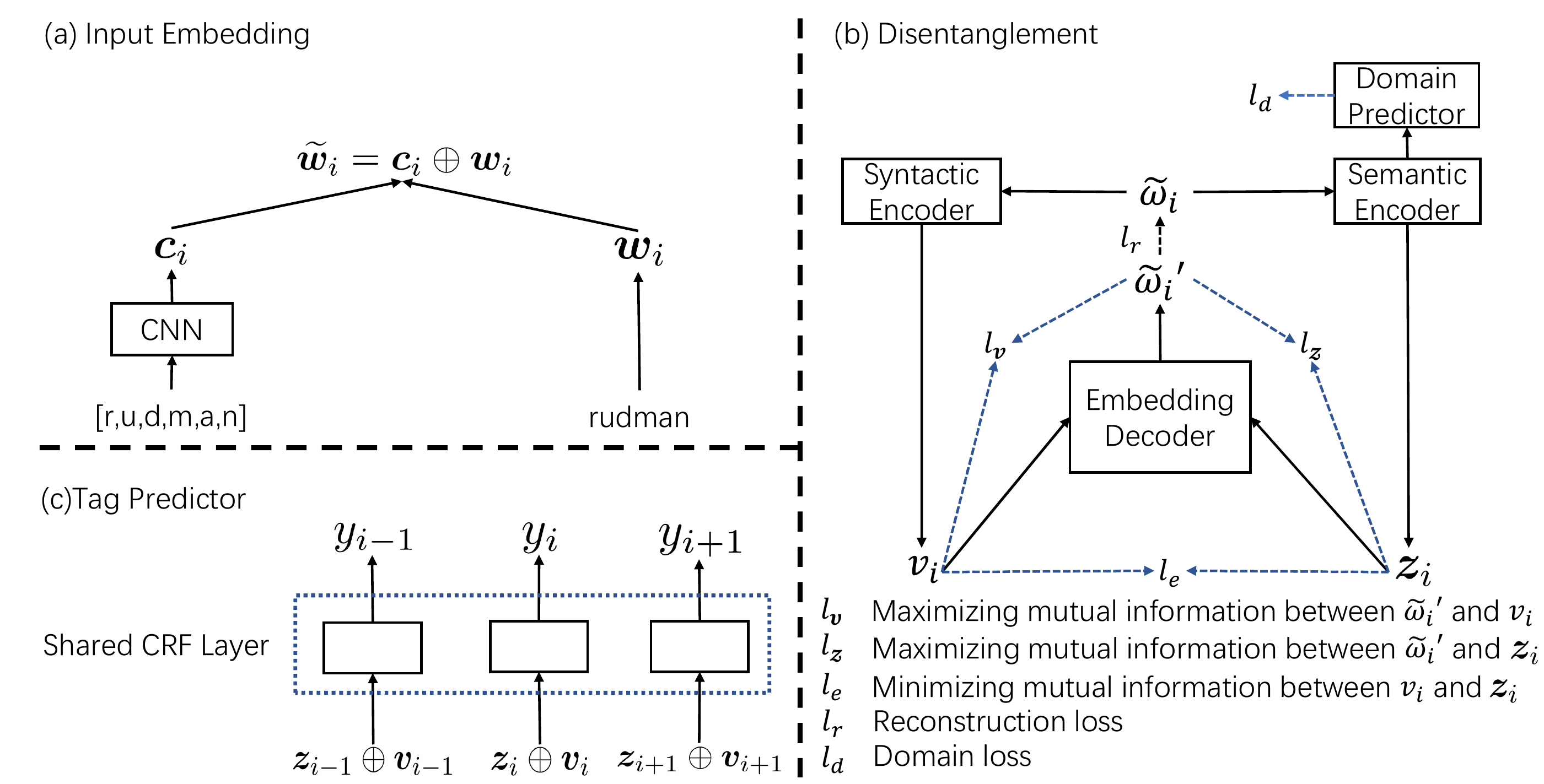}
	\caption{Architecture of the semi-supervised disentangled (\textbf{SSD}) framework for transferable NER. (a) The input embedding $\widetilde{\bm{\omega}}_i$ is generated by concatenating the word-level embedding $\bm{w}_i$ and character-level embedding $\bm{c}_i$. (b) The disentanglement between semantic latent variables $\bm{z}_i$ and syntactic structure latent variables $\bm{v}_i$. (c) The label $y$ is predicted by utilizing the disentangled semantic and syntactic structure latent variables simultaneously.}
	\label{fig:model}
\end{figure}

The framework of the proposed method is shown in Figure. \ref{fig:model} and it can be divided into three parts: input embedding with word-level and character-level information, mutual information regularization-based disentanglement, and tag prediction.

First, we generate the input embedding $\widetilde{\bm{w}}_i$
by concatenating the character-level embedding $\bm{c}_i$ and the word-level $\bm{\omega}_i$, and take it as the input for our model.

In contrast to the basic model that uses a single BiLSTM as the sequence model, we feed the input embedding into the semantic encoder $G_{\bm{z}}$ and the syntactic encoder $G_{\bm{v}}$ in order to obtain domain-specific latent variables $\bm{z}_i$ and domain-invariant latent variables $\bm{v}_i$.
It should be noted that the semantic encoder and structure encoder share the same architecture, i.e., a BiLSTM layer with a self-attention mechanism. Details of the BiLSTM and self-attention mechanism are given in Subsections \ref{bi-lstm} and \ref{self_attention}, respectively.
A decoder is used to reconstruct the input embedding $\widetilde{\bm{w}}_i'$ for each time step after receiving $\bm{z}_i$ and $\bm{v}_i$.
We concatenate $\bm{z}_i$ and $\bm{v}_i$ and feed them into the two-layer multi-layer perceptron (MLP) layers. The decoder is shared among all the time steps for the encoder output. Further details are provided in Subsection \ref{decoder}.

In order to disentangle the domain-specific latent variables $\bm{z}_i$ and domain-invariant latent variables $\bm{v}_i$, we use three mutual information regularization terms and domain label supervision. In particular, we minimize the mutual information between $\bm{z}_i$ and $\bm{v}_i$, and employ a domain predictor to determine whether the $\bm{z}_i$ comes from the source or target domain. Using the domain predictor, the domain-specific latent variables can be pushed into $\bm{z}_i$. By minimizing the mutual information between $\bm{z}_i$ and $\bm{v}_i$, we also make $\bm{z}_i$ and $\bm{v}_i$ independent. Subsequently, we further maximize the mutual information between $\bm{z}_i$ and $\widetilde{\bm{w}}_i'$ as well as the mutual information between $\bm{v}_i$ and $\widetilde{\bm{w}}_i'$.
Further details of the proposed SSD are given in the following sections.

\subsection{Latent Variable Reconstruction}\label{decoder}
In order to reconstruct the original input embedding, we employ the reconstruction architecture in the SSD framework, which contains a two-layer MLP and it is shared among all of the encoder time steps. Formally, we define the decoder as follows:
\begin{equation}
	\begin{split}
	\widetilde{\bm{w}_i}^{'}=MLP(MLP((\bm{z}_i\oplus\bm{v}_i))).
	\end{split}
\end{equation}

For convenience, we let $\bm{\psi}$ be the parameters of the decoder.
The loss function for the reconstruction is as follows:
\begin{equation}
	\begin{split}
	\mathcal{L}_r\left(\bm{\psi}, \Theta\right) = \frac{1}{L}\sum_{i=1}^{L} MSE(\widetilde{\bm{w}_i}^{'}, \widetilde{\bm{w}_i})
	\end{split}
\end{equation}

where $\bm{\psi}$ denotes the parameters of decoder and $MSE$ denotes the mean square error loss function.

\subsection{Domain-Specific Latent Variables Extraction} \label{domain_predictor}
In order to push the domain-specific information, such as topic information for different domains, into $\bm{z}_i$, we add a domain predictor $C_d$ that takes domain-specific latent variables $\bm{z}_i$ as the input and predicts the domain label. We use an MLP layer to predict the domain label for each sentence. Formally, this process can be defined as follows:
\begin{equation}
	\begin{split}
	d^* = MLP\left(Pooling\left(\bm{z}_1, \bm{z}_2, \cdots, \bm{z}_L;\bm{\theta_d}\right)\right),
	\end{split}
\end{equation}

where $Pooling(.)$ denotes the max pooling over all domain-specific latent variables at each time step. We use the cross entropy loss as the objective function for the domain predictor, as follows:
\begin{equation}
	\begin{split}
	\mathcal{L}_d = -d\log \left(d^*\right).
	\end{split}
\end{equation}

\subsection{ Regular Terms for Semi-supervised Disentanglement} \label{minesection}
\subsubsection{Mutual Information Neural Estimator}
In order to disentangle $\bm{z}$ and $\bm{v}$, we employ three mutual information regularization terms: minimizing the mutual information between $\bm{z_i}$ and $\bm{v}_i$, maximizing the mutual information between $\bm{z}_i$ and $\widetilde{\bm{w}}_i'$, and maximizing the mutual information between $\bm{v}_i$ and $\widetilde{\bm{w}}_i'$. Thus, the main challenge is finding a method that can estimate the mutual information between continuous random variables.
Fortunately, the mutual information neural estimator (MINE) \cite{belghazi2018mutual} can estimate the mutual information between latent variables using a neural network.
Formally, the mutual information between $A$ and $B$ can be described as follows:
\begin{equation}
\begin{split}\label{mi_bound_begin}
I(A,B):=H(A)-H(A|B),
\end{split}
\end{equation}

where $H$ is the Shannon entropy and $H(A|B)$ is the conditional entropy of $B$ given $A$. Furthermore, the mutual information is equivalent to the Kullback--Leibler divergence between the joint probability $\mathbb{P}\left(A,B\right)$ and the product of the marginals $\mathbb{P}\left(A\right) \otimes \mathbb{P}\left(B\right)$:
\begin{equation}\label{mine}
\begin{split}
I(A,B)=&D_{KL}\left(\mathbb{P}\left(A,B\right) || \mathbb{P}\left(A\right) \otimes \mathbb{P}\left(B\right) \right).
\end{split}
\end{equation}

In order to estimate the mutual information using a neural network, we follow the theorem proposed by Donsker et al.\cite{donsker1975asymptotic}.
\begin{theorem}
	(\textbf{Donsker--Varadhan representation.}\cite{donsker1975asymptotic}) The KL-divergence admits the following dual representation:
	\begin{equation}
	\begin{split}
	D_{KL}(\mathbb{P} || \mathbb{Q})= \mathop{\sup_{T:\Omega \rightarrow \mathbb{R}}} \mathbb{E}_{\mathbb{P}}\left[T\right]-\log\left(\mathbb{E}_{\mathbb{Q}}\left[e^T\right]\right),
	\end{split}
	\end{equation}
	where the supremum is taken over all functions $T$ such that the two expectations are finite.
\end{theorem}

In the equation above, $e^T$ is an exponential function, $T$ is a function that satisfies $\Omega \rightarrow \mathbb{R}$, and $\Omega$ denotes any function with finite integral. This theorem implies that we can estimate $D_{KL}(\mathbb{P} || \mathbb{Q})$ with a class of functions $T:\Omega \rightarrow \mathbb{R}$.

By combining this theorem with Equation (\ref{mine}), we obtain:
\begin{equation}
\begin{split}
I(A;B) &\geq I_{\phi}(A,B),
\end{split}
\end{equation}
where,
\begin{equation}\label{mi_bound}
\begin{split}
I_{\phi}(A,B)&=\mathbb{E}_{\mathbb{P}(A,B)}\left[T_\phi\right]-\log\left(\mathbb{E}_{\mathbb{P}\left(A\right) \otimes \mathbb{P}\left(B\right)}\left[e^{T_\phi}\right]\right),
\end{split}
\end{equation}
and function $T_\phi:\mathcal{A} \times \mathcal{B}\rightarrow \mathbb{R}$ is parameterized by a deep neural network with the parameter $\phi$. Therefore, we can estimate the mutual information between high dimensional continuous random variables by maximizing Equation (\ref{mi_bound}). We obtained Equations  (\ref{mi_bound_begin})--(\ref{mi_bound}) from the study by \cite{belghazi2018mutual}.

\subsubsection{Regularization Terms}
We extract the domain-specific latent variables with the domain predictor mentioned in \ref{domain_predictor}, but the domain-invariant information is entangled in a similar manner to the syntactic information. In order to address this problem, we employ three types of mutual information regularization terms for disentanglement. In particular, we can estimate the mutual information between $\bm{z}$ and $\bm{v}$ with the following method:
\begin{equation}
\begin{split}
I_{\phi_e}\left(\bm{z}_i, \bm{v}_i\right) \geq \mathbb{E}_{\mathbb{P}\left(\bm{z}_i, \bm{v}_i\right)}\left[T_{\phi_e}\right]-\log \left(\mathbb{E}_{\mathbb{P}\left(\bm{z}_i\right)\otimes\mathbb{P}\left(\bm{v}_i\right)}\left[e^{T_{\phi_e}}\right]\right),
\end{split}
\end{equation}
where $\phi_e$ is the parameter for $T_{\phi_e}$.

After using the domain predictor and minimizing the mutual information between $\bm{z}$ and $\bm{v}$, the model can disentangle the domain-invariant latent variables and the domain-specific latent variables. However, we must ensure that $\bm{v}$ contains the domain-invariant information. In the worst case, $\bm{v}$ may comprise the latent variables without any information. Therefore, we need to maximize the mutual information between $\widetilde{\bm{w}}_i'$ and $\bm{v}_i$, and maximize the mutual information between $\widetilde{\bm{w}}_i'$ and $\bm{z}_i$.

We estimate these two types of mutual information as follows:
\begin{equation}
\begin{split}
I_{\phi_{\bm{v}}}\left(\widetilde{\bm{w}_i}',\bm{v}_i\right) \geq \mathbb{E}_{\mathbb{P}\left(\widetilde{\bm{w}}_i', \bm{v}_i\right)}\left[T_{\phi_{\bm{v}}}\right] -
\log \left(\mathbb{E}_{\mathbb{P}\left(\widetilde{\bm{w}}_i'\right) \otimes \mathbb{P}\left(\bm{v}_i\right)}\left[e^{T_{\phi_{\bm{v}}}}\right]\right),
\end{split}
\end{equation}
\begin{equation}
\begin{split}
I_{\phi_{\bm{z}}}\left(\widetilde{\bm{w}_i}',\bm{z}_i\right) \geq \mathbb{E}_{\mathbb{P}\left(\widetilde{\bm{w}}_i', \bm{z}_i\right)}\left[T_{\phi_{\bm{z}}}\right] -
\log \left(\mathbb{E}_{\mathbb{P}\left(\widetilde{\bm{w}}_i'\right) \otimes \mathbb{P}\left(\bm{z}_i\right)}\left[e^{T_{\phi_{\bm{z}}}}\right]\right),
\end{split}
\end{equation}
where $\phi_{\bm{v}}$ and $\phi_{\bm{z}}$ are the parameters for $T_{\phi_{\bm{v}}}$ and $T_{\phi_{\bm{z}}}$ respectively. The objective functions can be defined as follows:
\begin{equation}
\begin{split}
&\mathcal{L}_{mi}\left(\phi_e, \phi_{\bm{z}}, \phi_{\bm{v}}\right) = l_{e}-l_{\bm{z}}-l_{\bm{v}},
\\&l_{e}\left(\phi_e\right)= \mathbb{E}_{\mathbb{P}\left(\bm{z}_i, \bm{v}_i\right)}\left[T_{\phi_e}\right]-\log, \left(\mathbb{E}_{\mathbb{P}\left(\bm{z}_i\right)\otimes\mathbb{P}\left(\bm{v}_i\right)}\left[e^{T_{\phi_e}}\right]\right) ,
\\&l_{\bm{v}}\left(\phi_{\bm{v}}\right)= \mathbb{E}_{\mathbb{P}\left(\widetilde{\bm{w}}_i', \bm{v}_i\right)}\left[T_{\phi_{\bm{v}}}\right] -
\log \left(\mathbb{E}_{\mathbb{P}\left(\widetilde{\bm{w}}_i'\right) \otimes \mathbb{P}\left(\bm{v}_i\right)}\left[e^{T_{\phi_{\bm{v}}}}\right]\right),
\\&l_{\bm{z}}\left(\phi_{\bm{z}}\right)= \mathbb{E}_{\mathbb{P}\left(\widetilde{\bm{w}}_i', \bm{z}_i\right)}\left[T_{\phi_{\bm{z}}}\right] -
\log \left(\mathbb{E}_{\mathbb{P}\left(\widetilde{\bm{w}}_i'\right) \otimes \mathbb{P}\left(\bm{z}_i\right)}\left[e^{T_{\phi_{\bm{z}}}}\right]\right).
\end{split}
\end{equation}

\subsection{Model Summary}
MINE \cite{belghazi2018mutual} can estimate the mutual information between two random variables with a certain distribution, but the distributions of $\bm{z}$ and $\bm{v}$ change when the model is trained because $\bm{z}$ and $\bm{v}$ are generated by the encoder. In practice, we implement the algorithm in an iterative training strategy. The formal procedure is presented in Algorithm 1.

\begin{algorithm}[H]
	\begin{algorithmic}[1]
		\Require $K_p, K_m, K_i, \eta$ are hyper-parameters. $K_p$ is the number of steps required to train the base model; $K_m$ is the number of steps required to train the mutual information neural estimator; $K_i$ is the number of steps for iterative training; $\eta$ is the learning rate.

		\For{$K_p$} \Comment{pre-train base model and domain predictor}
		\State $\Theta = \Theta - \eta\nabla\mathcal{L}_y$
		%		\State $\psi = \psi - \eta\nabla\mathcal{L}_r$
		\State $\theta_d = \theta_d - \eta \nabla \mathcal{L}_d$
		\EndFor
		%		\State
		\For{$K_i$}		\Comment{iterative training}
		\For{$K_m$}		\Comment{train mutual information estimator}
		\State $\Phi = \Phi - \eta \nabla \mathcal{L}_{mi}$
		\EndFor
		%		\State
		\For{$K_p$}		\Comment{disentanglement between $\bm{z}$ and $\bm{v}$}
		\State $\Theta = \Theta - \eta\nabla\mathcal{L}_y$
		\State $\psi = \psi - \eta\nabla\mathcal{L}_r$
		\State $\theta_d = \theta_d - \eta \nabla \mathcal{L}_d$
		\EndFor
		\EndFor
	\end{algorithmic}
	\label{aa}
	\caption{Minibatch stochastic gradient descent training for the SSD model. }
\end{algorithm}

In the training procedure, we employ the stochastic gradient descent algorithm to find the optimal parameters. In the prediction procedure, we input the target domain samples in the model and the labels of the target domain samples are predicted as follows:
\begin{equation}
\begin{split}
\hat{y_T}=G^T_y\left(G_a\left(G_r\left(G_e\left(x_T\right)\right)\right)\right).
\end{split}
\end{equation}

\section{Experimental} \label{experiment}
In the following, we introduce the data set employed for the evaluation and we then provide a brief introduction to the approaches compared. Finally, we present the experimental results.

\subsection{Data sets}
For cross-domain settings, the proposed approach was evaluated for four types of domains: \textit{Newswire}, \textit{Social Media}, \textit{Wiki},  and \textit{Spoken Queries}. For the Newswire domain, we used the OntoNotes 5.0 release data set (ON)\cite{weischedel2013ontonotes}. For the \textit{social media} domain, we employed the Ritter11 (R1) \cite{ritter2011named} data set. For the \textit{Wiki} domain, we employed the GUM \mbox{%DIFAUXCMD
\cite{zeldes2017gum} }\hspace{0pt}%DIFAUXCMD
data set. For the \textit{Spoken Queries} domain, we used the MIT Movie (MM) data set \cite{liu2013query}. Table \ref{dataset} shows details of the data sets from different domains. In contrast to the other baseline methods that purposely select a fixed source and target domains, we evaluated all of the methods across all of the transfer tasks. The statistics for these data sets are presented in Table \ref{cross_domain_statistic}.

We also evaluated our approach in cross-lingual settings for three different languages comprising \textit{Spanish} (S), \textit{Dutch} (D), and \textit{English} (E). For \textit{Spanish} and \textit{Dutch}, we used the \textbf{CoNLL-2002} data set \cite{tjong-kim-sang-2002-introduction}. For \textit{English}, we used the \textbf{CoNLL-2003} data set \cite{tjong2003introduction}. Furthermore, these data sets belong to the same domain and they share the same-named entity set. It should be noted that all three languages are Indo-European and they share the homologous syntactic structures. \textit{English} and \text{Dutch} belong to the Germanic group of languages, whereas \textit{Spanish} belongs to the Romance group of languages, so \textit{English} is closer to \textit{Dutch} and farther from \textit{Spanish}, and thus more homologous syntactic structures exist between \textit{Dutch} and \textit{English}. The statistics for these data sets are presented in Table \ref{cross_lingual_statistic}.

\begin{table}[H]
	\centering
	\caption{Named entities and their ratios in different data sets from different domains.}\label{dataset}
	\begin{tabular}{|l|l|p{10cm}|}
		\hline
		Name&Topic & Annotated Entities (\# ratio) \\ \hline \hline
		Ontonote-nw&\textit{Newswire} & Person (22\%), Location (2\%), Organization (38\%), NORP (8\%), GPE (26\%), Work of art (1\%), Event (0.8\%), Law (0.6\%), Facility (0.9\%), Product (1\%), Language (0.1\%) \\\hline
		Ritter2011&\textit{Social Media} & Person (30\%), \textit{Geo-loc} (18\%), Facility (6.9\%), Company (11\%), Sports Team (3.4\%), Music artist (3.6\%), Product (6.4\%), TV show (2.2\%), Movie (2.2\%), Other (15\%) \\\hline
		GUM&\textit{Wiki} & Abstract (24\%), Animal (1.5\%), Event (8\%), Object (12\%), \textit{Organization} (5\%), \textit{Person} (23\%), \textit{Place} (14\%), \textit{Plant} (1\%), Quantity (1\%), Substance (3\%), Time (4\%)  \\\hline
		Mit\_movie&\textit{Spoken Queries} & Actor (22\%), Character (5\%), Director (8\%), Genre (15\%), Plot (28\%), Year (14\%), Soundtrack (0.2\%), Opinion (4\%), Award (1.4\%), Origin (4\%), Quote (0.6\%), Relationship (3\%)  \\
		\hline
	\end{tabular}
\end{table}

\begin{table}[H]
	\begin{center}
		\centering
		\caption{Data set statistics for cross-domain setting.}\label{cross_domain_statistic}
		\begin{tabular}
			%			{|p{2.5cm}|p{2.5cm}|p{2.5cm}|p{2.5cm}|p{2.5cm}|}
			{|c|c|p{3cm}|c|c|}
			\hline
			Data set   & Language & \#Training Tokens & \#Dev Tokens & \#Test Tokens\\ \hline \hline
			Ontonote-nw& English  & 848200           & 144319      &  49235\\ \hline
			Ritter2011 & English  & 37098            & 4461        &  4730 \\ \hline
			Mit\_movie & English  & 158823           & -           &  39035\\ \hline
			GUM        & English  & 44111            & -           &  18236\\ \hline
		\end{tabular}
	\end{center}
\end{table}

\begin{table}[H]
	\begin{center}
		\centering
		\caption{Data set statistics for cross-lingual setting.} \label{cross_lingual_statistic}
		\begin{tabular}
			%			{|p{2.5cm}|p{2.5cm}|p{2.5cm}|p{2.5cm}|p{2.5cm}|}
			{|c|c|p{3cm}|c|c|}
			\hline
			Data set   & Language & \#Training Tokens & \#Dev Tokens & \#Test Tokens\\ \hline \hline
			CoNLL 2003 & English  & 204567           & 51578       &  49235\\ \hline
			CoNLL 2002 & Dutch    & 202932           & 37761       &  68994\\ \hline
			CoNLL 2002 & Spanish  & 207484           & 51645       &  52098\\ \hline
		\end{tabular}
	\end{center}
\end{table}

\subsection{Approaches Compared}
We compared the proposed SSD framework with the following baseline methods.
\begin{itemize}

	\item \textbf{In\_domain}: This method uses the limited target domain training data to train a model and applies this model to the test data without using the source domain data. This method does not transfer any knowledge from the source domain, so it is expected to provide the lower performance bound. It was also used as a baseline method by Lin et al.\cite{lin2018neural}.

	\item \textbf{Init\_tuning}: Init\_tuning is a straightforward method for transferable NER developed by Lee et. al \cite{lee2018transfer}. This method first trains a model using labeled source data and then treats it as the initialized model. This model is then fine tuned with the labeled data from the target domain.
	The output space for the target domain is different from that for the source domain, so the parameters of the target domain label predictor need to be updated by training with the target labeled data.

	\item  \textbf{Multi}: The multi-task-based method was developed by Yang et al. \cite{yang2016multi}. This method employs the idea of multi-task learning and it simultaneously trains two different classifiers by using the labeled source and target domain data. It should be noted that a feature extractor is shared between the source domain and target domain. In inference mode, we ignore the source classifier and obtain the predicted target label by feeding the target test data set.

	\item \textbf{Layer\_adaptation}:The Layer\_adaptation model \mbox{%DIFAUXCMD
\cite{lin2018neural} }\hspace{0pt}%DIFAUXCMD
was proposed by Lin et al. This method bridges the gap between heterogeneous input and output spaces by applying input and output adaptation layers. A pre-trained transferable word embedding is not available in the word adaptation layer, so we removed the word adaptation layer and used the same pre-trained word embedding without the word adaptation layer to ensure a fair comparison, and thus our analysis was orthometric.

	\item \textbf{Cross-Lingual Transfer Learning (CLTL)}: CLTL \mbox{%DIFAUXCMD
\cite{kim-etal-2017-cross} }\hspace{0pt}%DIFAUXCMD
is a learning model designed for part-of-speech tagging without ancillary resources. This cross-lingual model aims to extract common knowledge from other languages using a common BiLSTM and GRL \mbox{%DIFAUXCMD
\cite{ganin2015unsupervised}}\hspace{0pt}%DIFAUXCMD
, and a private BiLSTM for language-specific features. No restrictions are applied to  the language-specific BiLSTM, so this model cannot guarantee that the extracted feature is disentangled.
	\item \textbf{Multi-Task Cross-Lingual (MTCL)} Sequence Tagging Model: MTCL \mbox{%DIFAUXCMD
\cite{yang2016multi} }\hspace{0pt}%DIFAUXCMD
is a deep hierarchical recurrent neural network for sequence tagging. This model is similar to \textbf{Multi} but it simultaneously utilizes multiple languages.
\end{itemize}

Our model and the baseline methods were implemented with TensorFlow on a server with one GTX-1080 and Intel 7700K. To ensure fair comparisons, we applied the same hyper-parameter setting used by \mbox{%DIFAUXCMD
\cite{ma-hovy-2016-end} }\hspace{0pt}%DIFAUXCMD
for all of the methods. The hyper-parameters are shown in Table \ref{tb:exp:parachiller}.

\begin{table}[H]
		\centering
		\caption{Hyper-parameters used in all models.}\label{tb:exp:parachiller}
		\begin{center}
			\begin{tabular}{|c|c|}
				\hline
				Hyper-Parameter & Value \\
				\hline \hline
				Batch size & 64  \\ \hline
				Word embedding size & 100  \\\hline
				Char embedding size & 100 \\\hline
				Optimizer & Adam \\\hline
				Learning rate & 0.001 \\\hline
				Dropout rate & 0.5 \\
				\hline
			\end{tabular}
		\end{center}
\end{table}

\subsection{Results Based on Cross-domain Transfer}
We compared \textit{SSD} and the baseline methods using four different data sets in order: (1) to identify the factors that influence the performance of semi-supervised domain adaptation in NER, and (2) to assess the generality of our SSD model compared with other state-of-the-art approaches.
In order to answer these two questions, we quantitatively analyzed the experiment results.
To simulate conditions where labeled target domain data were unavailable, we randomly selected 10\% of the ON data set as the target domain data.

\subsubsection{Analysis of generalizability}
The experimental results also demonstrated the generalizability of our \textit{SSD} model. As shown in Table \ref{tab:office_home}, we found that our SSD model outperformed the other approaches in most of the transfer directions. For the transfer direction selected by many methods, i.e., ON$\rightarrow$R1, all of the approaches performed better with the in\_domain baseline, and our method achieved the best result. When we tested the reverse direction, i.e., R1$\rightarrow$ON, the other approaches lost their advantage because the proportions of common entities were different in R1 and ON. 
Initialization-based methods focus more on the domain-specific information in the target domain and they consider little of the domain-invariant information, whereas multi-task-based methods focus more on domain-invariant information and ignore the domain-specific information, so their performance is inferior. However, our \textbf{SSD} disentangles the domain-invariant and domain-specific information, and thus it can utilize both types of information to achieve better performance. For other transfer directions where the two domains were totally different, i.e., GUM and MM, our method still obtained comparable results. Thus, our method performed better than the baseline methods in all transfer directions and its generalizability was better.

\begin{table}[H]
	\centering
	\caption{F1-scores (\%) with four different domain data sets.} \label{cross_domain_result}
	\resizebox{15.5cm}{11.5mm}{
		\begin{tabular}{p{2.6cm}|ccccccccccccc}
			\hline
			 Method&R1$\rightarrow$ON&R1$\rightarrow$MM&R1$\rightarrow$GUM&ON$\rightarrow$R1&ON$\rightarrow$MM&ON$\rightarrow$GUM&MM$\rightarrow$ON&MM$\rightarrow$R1&MM$\rightarrow$GUM&GUM$\rightarrow$ON&GUM$\rightarrow$R1&GUM$\rightarrow$MM&Avg \\
			\hline
			In\_domain            & \large{85.9} & \large{72.4} &\large{53.1} & \large{64.7} & \large{72.4} & \large{53.1} & \large{85.9} & \large{64.7} & \large{53.1} & \large{85.9} & \large{64.7} & \large{72.4} & \large{69.0} \\
			INIT\_tuning          & \large{85.3} & \large{72.6} & \large{53.1} & \large{65.3} & \large{72.5} & \large{53.3} & \large{85.3} & \large{64.5} & \large{53.0} & \large{85.7} & \large{62.2} & \large{71.7} & \large{68.7} \\
			Layer\_adaption       & \large{85.3} & \large{72.6} & \large{53.1} & \large{65.3} & \large{72.5} & \large{53.3} & \large{85.3} & \large{64.5} & \large{53.0} & \large{85.7} & \large{62.2} & \large{71.7} & \large{68.7} \\
			Multi                 & \large{85.3} & \large{72.7} & \large{53.2} & \large{66.8} & \large{72.7} & \large{53.5} & \large{85.6} & \large{66.9} & \large{53.7} & \large{85.5} & \large{66.5} & \large{72.6} & \large{69.6} \\
			SSD                   & \large{\textbf{86.4}} & \large{\textbf{72.9}} & \large{\textbf{54.4}} & \large{\textbf{69.1}} & \large{\textbf{73.2}} & \large{\textbf{54.7}} & \large{\textbf{86.3}} &\large{\textbf{68.5}} & \large{\textbf{54.1}} & \large{\textbf{85.7}} & \large{\textbf{68.5}} & \large{\textbf{72.8}} & \large{\textbf{70.6}} \\
			\hline
	\end{tabular}}
	\label{tab:office_home}
\end{table}

\subsubsection{Analysis of the influence of semantic similarity}
The assumption that the target domain and source domain contain many common entities is usually excessively strong. In most cases, the entities in two domains are simply homogeneous and they share similar or related meanings, such as ``movie'' and ``TV show'' in the Rittter2011 domain, and ``actor'' and ``director'' in the Mit\_movie domain. In this case, the entities might be totally different in the source and target domains, but they share similar topics and can also be transferable.

According to Table \ref{dataset}, we found that the meanings of some entities in R1 were strongly related to those in MM, e.g., ``movie'' and ``TV show'' in R1 were related to ``actor'' and ``director'' in MM. In many semi-supervised transfer methods, R1$\rightarrow$MM and MM$\rightarrow$R1 perform better than GUM$\rightarrow$MM and MM$\rightarrow$GUM, which is also consistent with our assumption.
\begin{table}[H]
		\footnotesize
		\centering
		\caption{Statistics for common entities.}\label{common}
		\begin{tabular}{|c|p{12cm}|}
			\hline
			Name& Common Entities (Percentage of Source, Percentage of Target) \\ \hline \hline
			ON R1 &(Person (22\%), Person (30\%)), (Facility (0.9\%), Facility (6.9\%)), (Product (1\%), Product (6.4\%))\\\hline
			\hline
			ON GUM & (Person (22\%), Person (23\%)), (Organization (38\%), Organization (5\%)), (Event (0.8\%), Event (8\%))\\\hline
			ON MM &Null \\\hline
			GUM R1 &(Person (23\%), Person (30\%))\\\hline
			GUM MM &Null \\\hline
			R1 MM &Null \\\hline
		\end{tabular}
\end{table}

\subsubsection{Analysis of Disentangled Representation}
Intuitively, the amounts of common entity types in the source and target domains will influence the transferability. Thus, knowledge can be transferred more readily when there are more common entity types in the source and target domains.
The statistic for the common entities in different domains are presented in Table \ref{common}.

In order to evaluate the effectiveness of disentangled domain-invariant representation, we compared our \textbf{SSD} model with In\_domain and Multi based on the common entities, which are shown as the common entities in Table \ref{tab:common_and_non_common}. As mentioned above, the Multi method treats each classification from a different domain as a task and aims to extract the representation that is shared between tasks, so the performance of Multi exceeded that of In\_domain in most tasks. However, this method cannot avoid the influence of negative transfer from the non-common entities because the representation extracted by Multi is distorted on the feature manifold. This is why Multi performed worse than In\_domain in some tasks, e.g., R1 $\rightarrow$ ON, MM $\rightarrow$ ON, and GUM $\rightarrow$ ON. However, our \textbf{SSD} method disentangles the domain-invariant and domain-specific information to avoid this problem, and thus it performed better than In\_domain in all tasks. It should be noted that the improvement obtained with our \textbf{SSD} method was not as remarkable when the target domain was ON because ON is easy to train, and we obtained a very high f1 score (more than 85\%) in the In\_domain setting.

In contrast to Multi, our \textbf{SSD} method also utilizes the domain-specific information. In order to study the effectiveness of disentangled domain-specific representation, we compared our \textbf{SSD} model with In\_domain and Multi for the non-common entities, and the non-common entity results are shown in Table \ref{tab:common_and_non_common}. Multi does not explicitly utilize the domain-specific representation, so the performance of Multi was worse than that of In\_domain, e.g.,  R1 $\rightarrow$ ON and MM $\rightarrow$ ON. However, our \textbf{SSD} method uses  both the domain-invariant and domain-specific information at the same time, so \textbf{SSD} performed better than the baseline methods at most tasks. However, we also found that the performance declined when we employed GUM as the source domain because GUM contained some incorrectly labeled entities.

\begin{table}
	\centering
	\footnotesize
	\begin{center}
		\caption{F1-scores (\%) for common and non-common entities.}
		\label{tab:common_and_non_common}
		\begin{tabular}{|p{2.0cm}|p{1.5cm}|p{1.7cm}|p{1.7cm}|p{1.5cm}|p{1.7cm}|p{1.7cm}|}
			\hline
			&\multicolumn{3}{|c|}{Common entity results}&\multicolumn{3}{|c|}{Non-common entity results} \\ \hline
			Transfer task        & in\_domain& Multi                  &   SSD                 & in\_domain & Multi                  &SSD                     \\ \hline
			{R1$\rightarrow $ON} &49.24      &48.93 (0.49$\downarrow$) &49.92 (0.68$\uparrow$)  &36.61       & 36.38 (0.23$\downarrow$)&36.83 (0.22$\uparrow$)   \\ \hline		
			{R1$\rightarrow $MM} &24.63      &24.70 (0.07$\uparrow$)   &24.85 (0.23$\uparrow$)  &48.20       & 48.32 (0.12$\uparrow$)  &48.51 (0.31$\uparrow$)   \\ \hline	
			{R1$\rightarrow $GUM}&29.42      &29.49 (0.07$\uparrow$)   &30.33 (0.91$\uparrow$)  &23.77       & 24.10 (0.33$\uparrow$)  &24.27 (0.50$\uparrow$)   \\ \hline	
			{ON$\rightarrow $R1} &54.55      &55.47 (0.92$\uparrow$)   &56.94 (2.39$\uparrow$)  &10.22       & 10.42 (0.20$\uparrow$)  &10.94 (0.72$\uparrow$)   \\ \hline	
			{ON$\rightarrow $MM} &45.48      &45.55 (0.07$\uparrow$)   &45.95 (0.47$\uparrow$)  &27.34       & 27.41 (0.07$\uparrow$)  &27.56 (0.22$\uparrow$)   \\ \hline	
			{ON$\rightarrow $GUM}&32.49      &32.54 (0.05$\uparrow$)   &33.53 (1.04$\uparrow$)  &20.70       & 20.82 (0.12$\uparrow$)  &21.16 (0.46$\uparrow$)   \\ \hline	
			{MM$\rightarrow $ON} &14.91      &14.69 (0.22$\downarrow$) &14.95 (0.04$\uparrow$)  &70.90       & 70.57 (0.33$\downarrow$)&70.96 (0.06$\uparrow$)   \\ \hline	
			{MM$\rightarrow $R1} &25.25      &25.67 (0.42$\uparrow$)   &26.57 (1.32$\uparrow$)  &39.51       & 40.21 (0.70$\uparrow$)  &40.45 (0.94$\uparrow$)   \\ \hline	
			{MM$\rightarrow $}GUM&24.19      &24.41 (0.22$\uparrow$)   &25.04 (0.85$\uparrow$)  &29.00       & 29.47 (0.47$\uparrow$)  &29.54 (0.54$\uparrow$)   \\ \hline	
			{GUM$\rightarrow $ON}&49.26      &48.95 (0.31$\downarrow$) &49.32 (0.07$\uparrow$)  &36.59       & 36.47 (0.12$\downarrow$)&36.28 (0.31$\downarrow$) \\ \hline	
			{GUM$\rightarrow $R1}&51.62      &52.52 (0.90$\uparrow$)   &53.36 (1.74$\uparrow$)  &13.14       & 12.65 (0.49$\downarrow$)&12.86 (0.28$\downarrow$) \\ \hline	
			{GUM$\rightarrow $MM}&45.48      &45.64 (0.16$\uparrow$)   &45.80 (0.32$\uparrow$)  &27.34       & 27.25 (0.09$\downarrow$)&27.16 (0.18$\downarrow$) \\ \hline	
		\end{tabular}
	\end{center}
\end{table}

\subsection{Results Based on CLTL}
We compared the performance of \textbf{SSD} and the other methods with three different language data sets derived from \textbf{CoNLL-2002} and \textbf{CoNLL-2003}. These data sets contained three different languages comprising \textit{Spanish}, \textit{Dutch}, and \textit{English}, and they were all related to the same topic (i.e., news). The syntactic structure of \textit{English} is similar to that of the other two languages to some extent. Both \textit{Dutch} and \textit{English} belong to the Germanic group of language, whereas \textit{Spanish} belongs to the Romance group of language. Thus, \textit{Dutch} is similar to \textit{English}, whereas \textit{Spanish} is not. In the cross-lingual transfer experiment, we assumed that the syntactic structure was domain specific and the semantics were domain invariant. Therefore, we used \textbf{SSD} to disentangle the domain-invariant semantics and domain-specific syntactic structure, before finally applying both for CLTL. In order to evaluate the effectiveness of our model and to consider the  semi-supervised domain adaptation setting, we randomly selected 20\% of the data as each target domain training data set.

\begin{table}[H]
	\centering
	\caption{F1-scores (\%) with three different language data sets.}
	\begin{tabular}{|l|ccccccc|}
		\hline
		Methods         & $E \rightarrow S$ & $E \rightarrow D$ & $S \rightarrow E$ & $S \rightarrow D$ & $D \rightarrow E$ & $D \rightarrow S$ & Avg\\
		\hline
		In\_domain            & 71.3 & 63.2 & 75.7 & 63.2 & 75.7 & 71.2 & 70.5 \\
		Init\_transfer        & 71.7 & 66.5 & 76.0 & 64.2 & 76.2 & 71.5 & 71.0 \\
		Multi\_transfer       & 71.7 & 65.1 & 76.2 & 64.8 & 75.3 & 71.6 & 70.9 \\
		MTCL                  & 72.1 & 67.1 & 76.3 & 67.1 & 76.3 & 72.1 & 71.8 \\
		CLTL                  & 73.6 & 67.2 & 76.9 & 67.7 & 77.0 & 74.3 & 72.8 \\
		SSD                   & \textbf{75.0} & \textbf{69.8} & \textbf{80.9} & \textbf{68.7} & \textbf{81.0} & \textbf{74.7} & \textbf{75.0} \\
		\hline
	\end{tabular}

	\label{tab:office_31}
\end{table}

As shown in Table \ref{tab:office_31}, \textbf{SSD} performed better than the other methods in all transfer directions and the improvement in the F1-score was quite impressive. For some transfer directions, SSD achieved improvements of more than four points compared with Multi and Init, thereby indicating that both domain-invariant and domain-specific information are important, and the \textbf{SSD} model could disentangle and capture this information, before finally utilizing it to obtain better predictions.

\subsection{Low-resource Corpora Setting}
\begin{figure}[H]\centering
	\centering

	\subfigure[ON$\rightarrow$R1]{
		\includegraphics[width=0.47\textwidth]{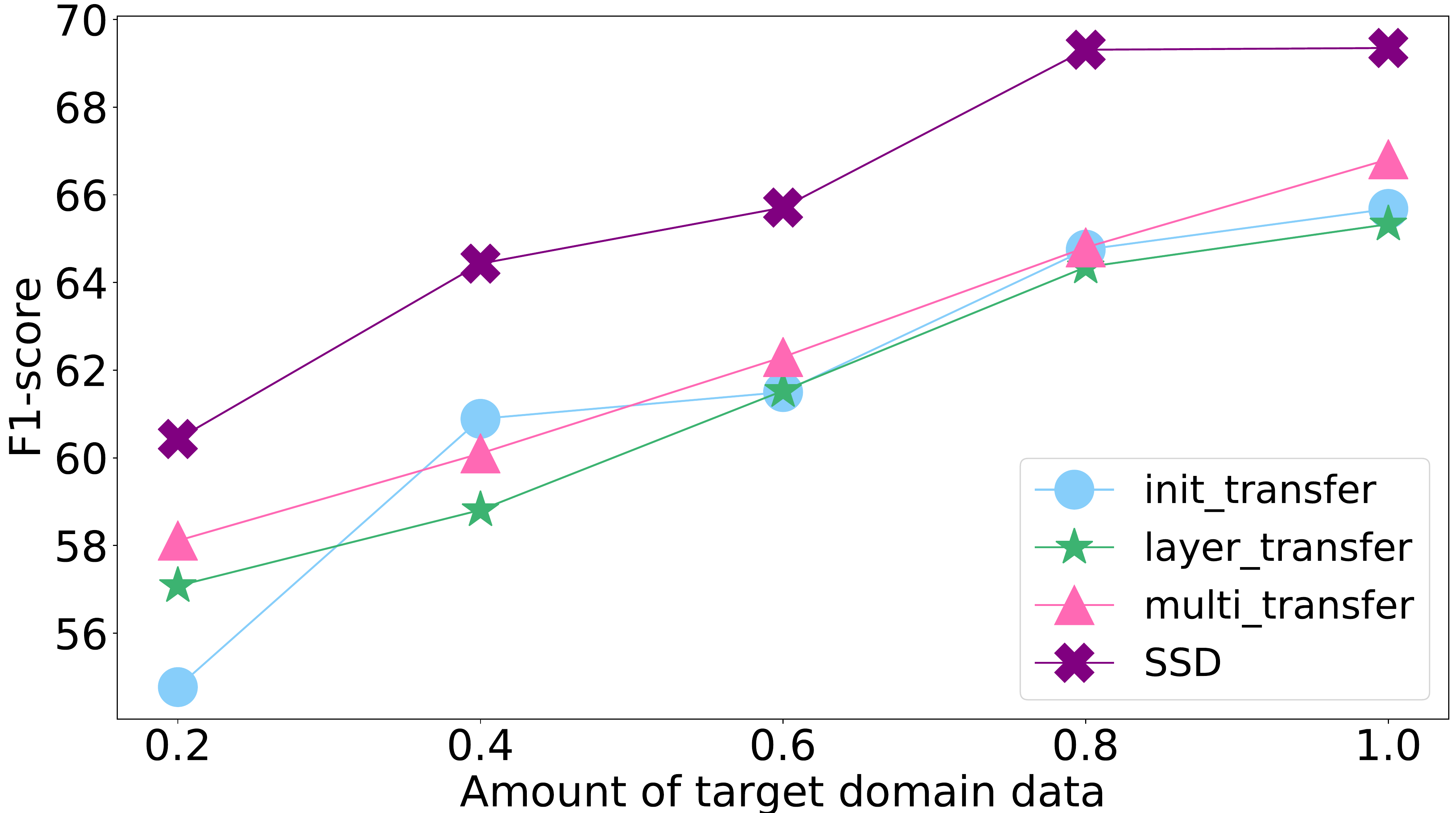}
		\label{fig:ON2R1}
	}
	\subfigure[ON$\rightarrow$GUM]{
		\includegraphics[width=0.47\textwidth]{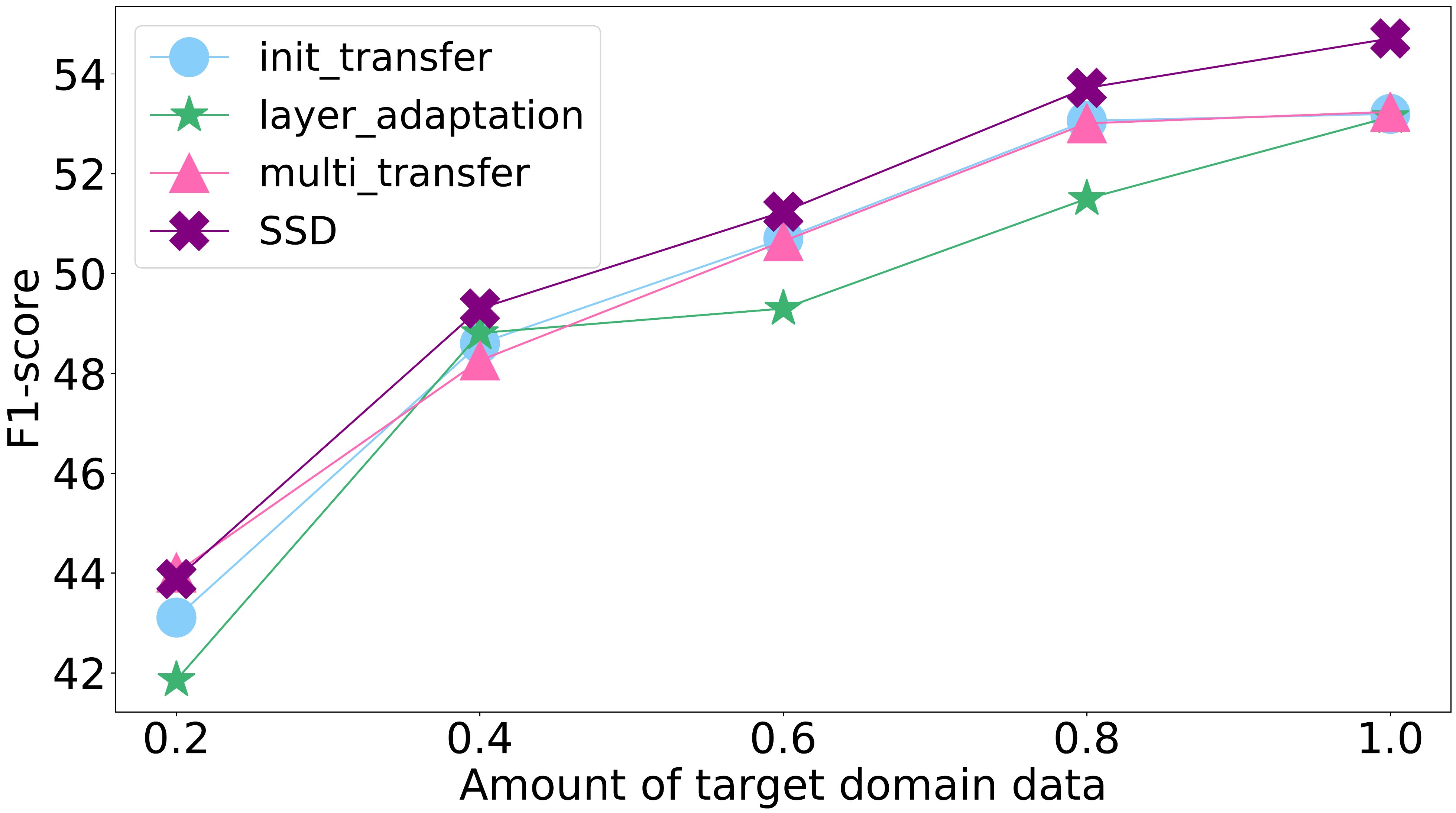}
		\label{fig:ON2GUM}
	}
	\subfigure[ON$\rightarrow$MM]{
		\includegraphics[width=0.47\textwidth]{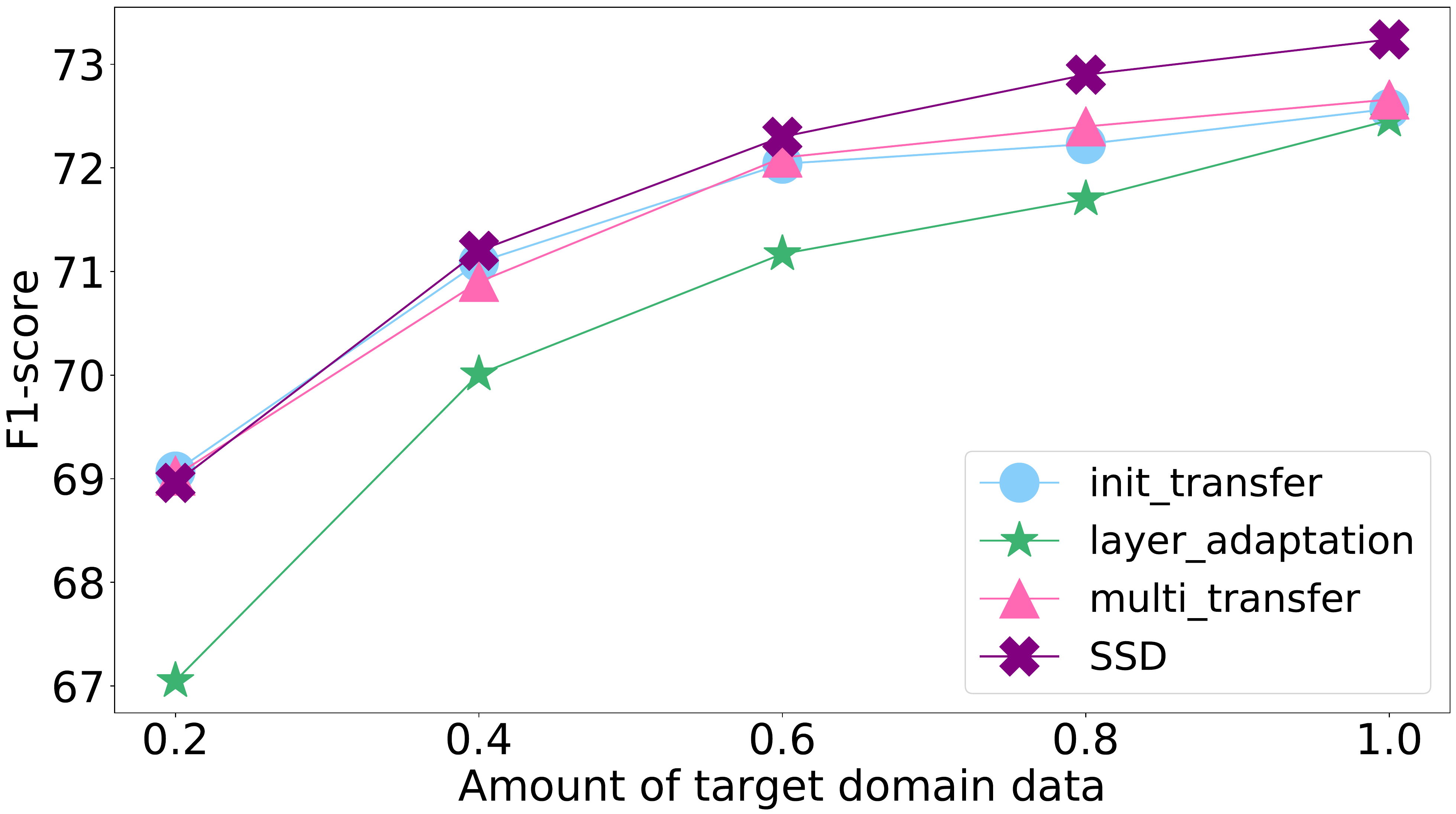}
		\label{fig:ON2MM}
	}
	\subfigure[\textit{English}$\rightarrow$\textit{Dutch}]{
		\includegraphics[width=0.47\textwidth]{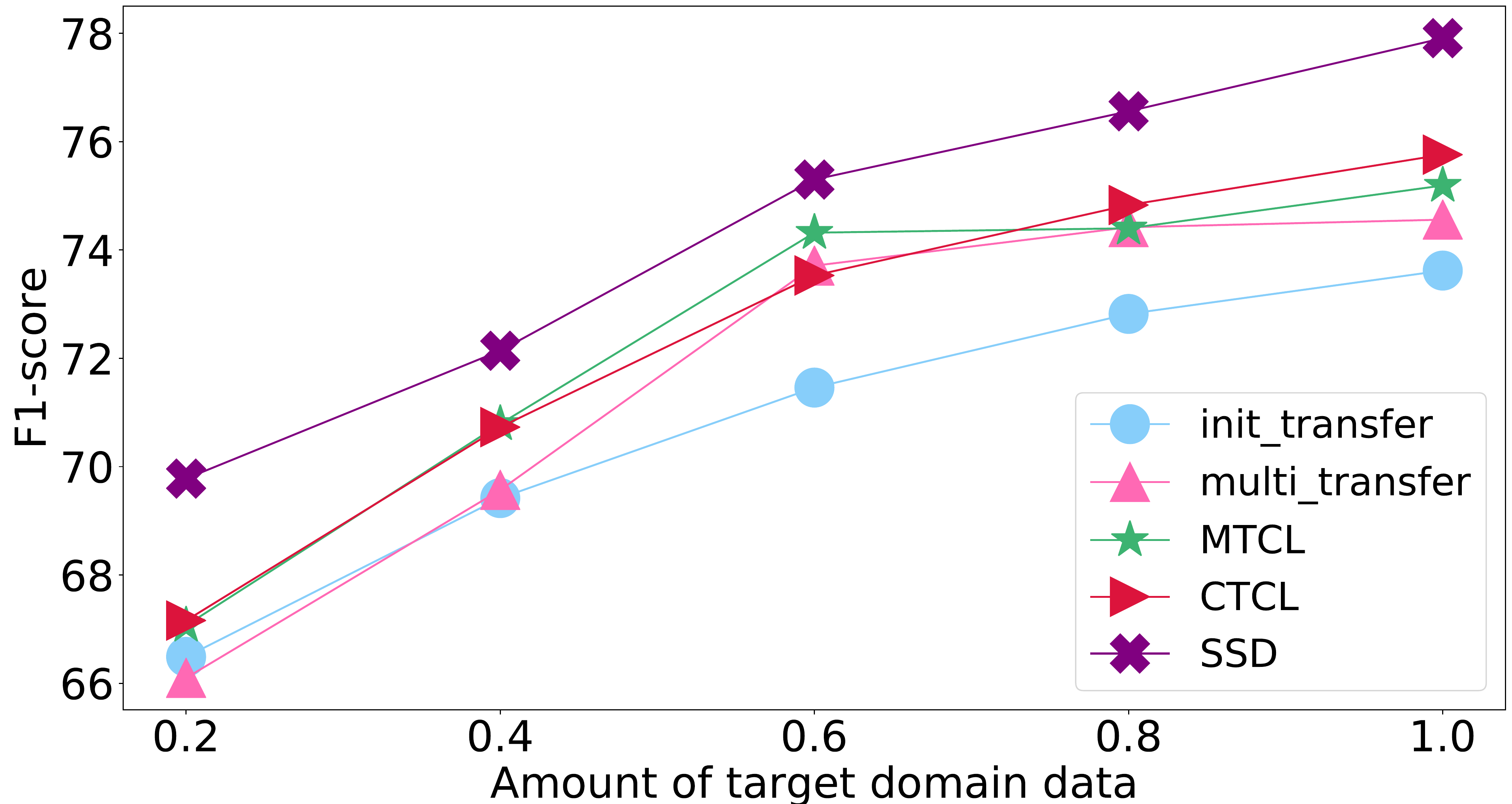}
		\label{fig:E2D}
	}
	\subfigure[\textit{English}$\rightarrow$\textit{Spanish}]{
		\includegraphics[width=0.47\textwidth]{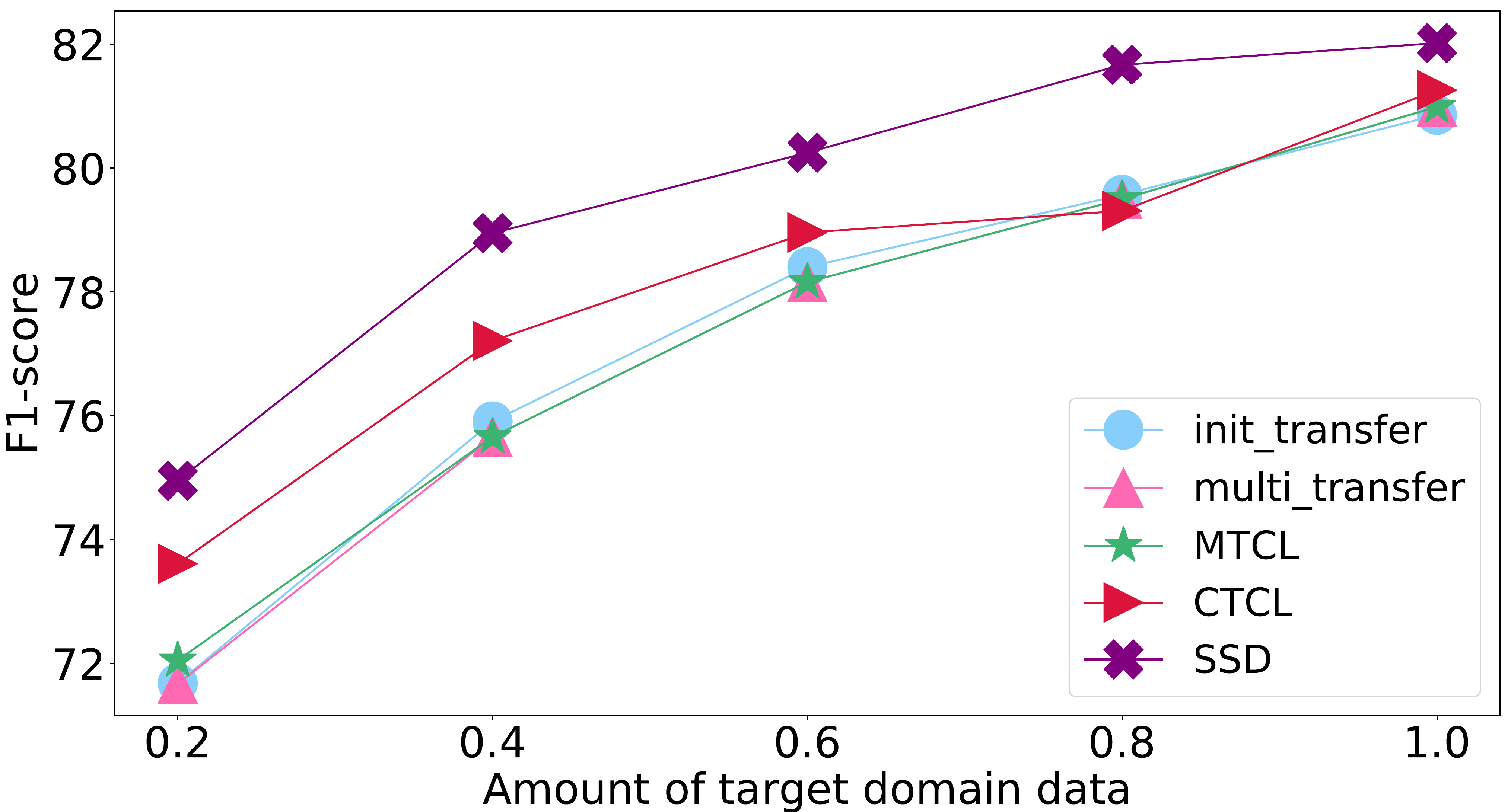}
		\label{fig:E2S}
	}
	\vskip 4pt
	\caption{(a) F1-score vs. amount of target domain data in ON $\rightarrow$ R1 transfer direction. (b) F1-score vs. amount of target domain data in ON $\rightarrow$ GUM transfer direction. (c) F1-score vs. amount of target domain data in ON $\rightarrow$ MM transfer direction. (d) F1-score vs. amount of target domain data in \textit{English}$\rightarrow$\textit{Dutch} transfer direction. (e) F1-score vs. amount of target domain data in \textit{English}$\rightarrow$\textit{Spanish} transfer direction.}
	\label{fig:toys}
\end{figure}

In order to assess the effectiveness of our approach when the amounts of training data from the target domain were limited, we conducted further experiments by gradually increasing the size of the target training data from 20\% to 100\%. Fig. \ref{fig:ON2R1}, Fig. \ref{fig:ON2GUM}, and Fig. \ref{fig:ON2MM} illustrate the cross-domain experimental results for ON$\rightarrow$R1, ON$\rightarrow$GUM, and ON$\rightarrow$MM, respectively. In addition, Fig. \ref{fig:E2D} and Fig. \ref{fig:E2S} illustrate the cross-lingual experiment results for \textit{English} to \textit{Dutch} and \textit{English} to \textit{Spanish}, respectively.

\subsubsection{Low-resource Cross-domain Setting}
In a low-resource cross-domain setting, we also found that when the proportion of the target domain training data was small, all methods failed to achieve ideal performance, but our SSD model still obtained comparable results. As the size of the target data set increased, the difference between our model and the other baseline methods increased because more domain-specific information was available, thereby improving the performance of our model. When the scale of the target domain data set varied, our SSD model consistently achieved the best results, thereby demonstrating that: (1) domain-invariant and domain-specific information both contributed to the performance of transferable NER; (2) the performance of the multi-task-based methods increased slowly when the size of the target domain data set was large because these methods only focus on extracting the domain-invariant information from both domains, whereas they ignore the domain-specific information to some extent; and (3) our SSD model disentangled the domain-invariant and domain-specific information, and utilized both simultaneously to achieve the best results.

\subsubsection{Low-resource Cross-lingual Setting}
In the low-resource cross-lingual setting, both the source and target domains belonged to the same domain. As the size of the target domain data set increased, the performance of all methods improved. In contrast to the results obtained in the low-resource cross-domain setting, we also found that when the size of the target domain data set was small, our method still obtained ideal performance, whereas the performance of the other methods decreased rapidly. When the amount of target training data was small, the initialization-based methods may have been affected by overfitting and the multi-task-based method could only extract a small amount of common information. By contrast, our SSD method explicitly captured the domain-invariant information and utilized the domain-specific information in the target domain to guarantee better performance.

\subsection{Ablation Study}
To further investigate the effectiveness of each component of the model, we compared \textbf{SSD} with the following variants.
\begin{itemize}
	\item \textbf{SSD-nAttn}: No attention mechanism in the \textbf{SSD} model.
	\item \textbf{Simple-Attn}: We remove the disentanglement mechanism from the \textbf{SSD} model and the model degenerated to the simple Char-LSTM + attention model.
	\item \textbf{SSD-RD}: To study the quality of disentanglement, we removed the objective function for disentangling these two latent variables in the \textbf{SSD} model.
	\item \textbf{SSD-DS}: To assess whether the domain-specific information could  improve the performance of the model compared with the multi-task-based method, we also tested \textbf{SSD-DS} where the domain-invariant encoder and decoder were  removed. In this case, some domain-specific information was considered because of the domain predictor.
\end{itemize}

\begin{table}[H]
		\centering
		\caption{Evaluation of different SSD components.}\label{tab:ablation}
		\begin{tabular}{|l|cccccc|}
			\hline
			Methods           & $E \rightarrow S$ & $E \rightarrow D$ & $S \rightarrow E$ & $S \rightarrow D$ & $D \rightarrow E$ & $D \rightarrow S$\\
			\hline
			In\_domain        & 71.3 & 63.2 & 75.7 & 63.2 & 75.7 & 71.3\\
			Multi\_transfer   & 71.7 & 66.1 & 76.2 & 64.8 & 75.3 & 71.6\\
			Simple-Attn       & 73.5 & 67.6 & 77.1 & 67.3 & 77.3 & 73.7 \\
			%		Multi                 & 71.7 & 72.4 & 53.1 & 64.7 & 72.4 & 53.1 \\
			SSD-nAttn         & 73.2 & 66.8 & 76.8 & 66.6 & 76.8 & 72.9 \\
			SSD-DS            & 74.1 & 68.6 & 78.5 & 67.8 & 78.2 & 74.3 \\
			SSD-RD            & 74.5 & 69.2 & 79.1 & 68.2 & 79.6 & 74.5 \\
			SSD               & \textbf{75.0} & \textbf{69.8} & \textbf{80.9} & \textbf{68.7} & \textbf{81.0} & \textbf{74.7}\\
			%		$\delta$                & 3    & 4.6  & 4.7  & 4    & 4.7  & 3 \\
			\hline
		\end{tabular}
\end{table}

The results of the ablation study are shown in Table \ref{tab:ablation}. We found  that both the attention mechanism and SSD component considerably affected the model’s  performance. Furthermore, we observed the following.
1) The combination of the SSD component and syntactic-extraction attention mechanism, i.e., \textbf{SSD}, obtained superior performance compared with each individual component, thereby demonstrating their importance and complementary effect.
2) The model without the disentanglement mechanism (\textbf{Simple-Attn}) also performed better than \textit{Multi} because the syntactic structure extracted by the attention mechanism improved the transfer capability.
3) Compared with the standard \textbf{SSD}, the performance of \textbf{SSD-RD} was lower, which indicates that the disentanglement of the domain-invariant and domain-specific latent variables contributed to the improved performance. We also found that the entangled domain-specific representation could lead to negative transfer.
4) In order to study the effectiveness of domain-specific information, we also investigated \textbf{SSD-DS}, where the multi-task-based model used the domain predictor in order to preserve the domain-invariant information. In contrast to Multi\_transfer, the \textbf{SSD-DS} variant also utilized the domain-specific information and it obtained better results. However, it performed worse than the standard \textbf{SSD}, thereby demonstrating the effectiveness of disentanglement.

\section{Conclusion} \label{conclusion}
In this study, we proposed a novel SSD framework for NER. By deploying this SSD framework, we can successfully disentangle domain-invariant and domain-specific information with limited supervision of the signals. We demonstrated the usefulness of the proposed approach in cross-domain and cross-lingual settings. The results obtained in both cases showed the positive benefits of our method even when the size of the target domain was very small and it achieved state-of-the-art performance in the NER task. The success of the proposed SSD demonstrates that the disentanglement framework is an effective solution for domain adaptation tasks. In future research, this disentanglement framework can be extended to various NLP tasks.

\section*{Acknowledgments}
This study was supported partly by the NSFC-Guangdong Joint Fund (U1501254), Natural Science Foundation of China (61876043), Natural Science Foundation of Guangdong (2014A030306004, 2014A030308008), Guangdong High-level Personnel of Special Support Program (2015TQ01X140), and Science and Technology Planning Project of Guangzhou (201902010058).



	%%  \bibitem{key}...
	%%


	% \end{thebibliography}

\end{document}